%% file: DiffusionRG-ICLR.tex
\documentclass{article}
\usepackage{iclr2025_conference,times}

\usepackage{tikz}
\usetikzlibrary{calc}

\input{math_commands.tex}

\usepackage{ctable,multirow}
\usepackage{hyperref}
\usepackage{url}
\usepackage{physics}
\usepackage{graphicx}
\usepackage{wrapfig}
\usepackage{float}

\newcommand{\RR}{{\mathbb R}}

\newcommand{\bbe}{{\boldsymbol{\beta}}}
\newcommand{\bphi}{{\boldsymbol{\phi}}}
\newcommand{\bchi}{{\boldsymbol{\chi}}}

\title{GUD: Generation with Unified Diffusion}

\author{Mathis Gerdes\thanks{Correspondence: \texttt{m.gerdes@uva.nl}} \\
Institute of Physics \\
University of Amsterdam, NL \\
\And
Max Welling \\
Amsterdam Machine Learning Lab \\
University of Amsterdam, NL
\And
Miranda C. N. Cheng\thanks{On leave from CNRS, France.} \\
Institute of Physics, University of Amsterdam, NL \\
Institute for Mathematics, Academia Sinica, Taiwan \\
Korteweg-de Vries Institute for Mathematics, University of Amsterdam, NL \\
}

\iclrfinalcopy
\begin{document}

\maketitle

\begin{abstract}

Diffusion generative models transform noise into data by inverting a process that progressively adds noise to data samples. Inspired by concepts from the renormalization group in physics, which analyzes systems across different scales, we revisit diffusion models by exploring three key design aspects: 1) the choice of representation in which the diffusion process operates (e.g. pixel-, PCA-, Fourier-,\linebreak or wavelet-basis), 2) the prior distribution that data is transformed into during diffusion (e.g. Gaussian with covariance $\Sigma$), and 3) the scheduling of noise levels applied separately to different parts of the data, captured by a component-wise noise schedule.
Incorporating the flexibility in these choices, we develop a unified framework for diffusion generative models with greatly enhanced design freedom. In particular, we introduce soft-conditioning models that smoothly interpolate between standard diffusion models and autoregressive models (in any basis), conceptually bridging these two approaches.
Our framework opens up a wide design space which may lead to more efficient training and data generation, and paves the way to novel architectures integrating different generative approaches and generation tasks.

\end{abstract}

\section{Introduction}\label{intro}

Diffusion-based generative models, first introduced in \citet{sohl-dickstein2015DeepUnsupervisedLearning}, have seen great successes in recent years since the works of \citet{song2020GenerativeModelingEstimating,ho2020DenoisingDiffusionProbabilistic}.
In these models, data are transformed into noise following a diffusion process, and a transformation simulating the reverse process is learned which is then used to map noise into generated samples.
In physics, the theory of renormalization group (RG) flows has been a basic tool in the study of a wide range of physical phenomena, including phase transitions and fundamental physics, both in theoretical as well as numerical approaches.
In short, an RG flow prescribes a way of erasing the high-frequency information of a physical theory, while retaining the information relevant for the long-wavelength physics.
As such, there are clear analogs between score-based generative models and RG flows, at least at a conceptual level. Indeed, it has been known for a long time that RG flows in quantum field theories can also be described as a diffusive process \citep{Zinn-Justin:2002ecy,doi:10.1142/S0217751X01004670,cotler2023RenormalizingDiffusion, berman2023BayesianRenormalization}.
In both cases, information gets erased along the flow and many different initial distributions get mapped into the same final distribution -- a feature often referred to as ``universality" in the physics literature.
In the diffusion context, the ``universal'' distribution is given by the chosen noise distribution, independent of the data distribution.

However, there are salient differences between the ways diffusion models and RG erase information. First, the  {\bf{basis}}: the diffusive RG process is diagonal in the frequency-basis while the standard diffusion models typically diffuse diagonally in the pixel-basis. Second, the {\bf{prior distribution}}: the endpoint of RG is a scale-invariant distribution, often with the same second-order statistics as the distribution one starts with at the beginning of the RG flow. The standard diffusion models on the other hand indiscriminately map all data distributions to that of white noise.  Third, the {\bf{component-wise noising schedule}}: RG flows erase information sequentially from high to low frequencies, while the original diffusion model has the same noise schedule for all pixels.
In our chosen basis, we allow each component to have its own noising schedule.
These considerations lead us to the framework of generative unified diffusion (GUD) models which incorporate the freedom in design choices in the above three aspects.

Autoregressive models, such as next-token prediction models, play an increasingly dominant role in modern-day machine-learning applications such as LLMs, and seem to be distinct from diffusion models at first glance. In autoregressive models, tokens are generated one at a time, conditional on previously generated ones, while diffusion models generate information in all components simultaneously. We will show that the two can in fact be unified in our framework, which in particular allows for {\emph{soft-conditioning}} generative processes. Intuitively, this means that we can condition on partial information from other components as long as that information has already been generated in the diffusion process.

\section{Related Work}

The idea that diffusion models can incorporate the multi-scale nature of the dataset through a specific choice of data representation has been exploited in interesting works including Wavelet Score-Based Generative Models \citep{guth2022WaveletScoreBased}, Pyramidal Diffusion Models
\citep{ryu2022pyramidaldenoisingdiffusionprobabilistic}, Cascaded Diffusion Models \citep{ho2022CascadedDiffusionModels}, and more.
In these works, the generation is sharply autoregressive
 between different hierarchies of generation (between different resolutions, for instance).
 In Blurring Diffusion Models \citep{blurring}, diffusion models in the  frequency basis was proposed.
Our framework, besides not
being restricted to these specific choices of basis, allows for an extension of these work by introducing the freedom to soft-condition the generative process.

The possibility of interpolating between the standard diffusion models and token-wise autoregressive models has recently been explored in \citet{chen2024DiffusionForcing} in the context of causal sequence generation. In this work, a training paradigm where diffusion models, with a chosen token-wise noising schedule, are trained to denoise tokens based on the information of only the (partially noised) previous tokens, captured in latent variables. Our work integrates the choices of data representation, the prior distribution, and the component-wise noising schedule into one framework, and therefore includes this particular paradigm as a special case. In particular, as we demonstrate in experiments, it is possible to integrate multi-scale and spatially sequential generation processes in our framework.

In practice, for applications with high-dimensional data, the diffusion generation often takes place in a lower-dimensional latent space
\citep{rombach2022HighResolutionImage, sinha2021DiffusionDenoisingModels, vahdat2021scorebasedgenerativemodelinglatent}. The freedom to choose the basis proposed in our work is not to be understood as replacing latent space diffusion. Rather, our framework can straightforwardly be used in the latent space, leading to a latent GUD model.

\section{Preliminaries}
\subsection{Stochastic Differential Equations}
In continuous time, the general diffusion setup can be described by the following It\^o stochastic differential equation (SDE):
\begin{equation}\label{eqn:generalSDE}
    \dd{\bphi} = \mathbf{f}(\bphi, t) \dd{t} + \mathbf{G}(\bphi, t) \dd{\mathbf{w}}\,,
\end{equation}
where $\dd{\mathbf{w}}$ represents a white noise Wiener process.
We use $\bphi\in \RR^d$ to denote a vector.
In the above, we have $\mathbf{f}(\,\cdot\,, t): \RR^d\to \RR^d$ and $\mathbf{G}(\,\cdot\,, t) : \RR^d\to \RR^{d\times d}$.
The reverse-time SDE is given by \citep{anderson1982ReversetimeDiffusion}
\begin{equation}\label{eqn:generalinverseSDE}
  \dd{\bphi} = \left(\mathbf{f}(\bphi, t)
  -\nabla\cdot(\mathbf{G}\mathbf{G}^T) (\bphi, t) - \mathbf{G}\mathbf{G}^T \nabla_\bphi \log p_t(\bphi) \right) \dd{t} + \mathbf{G}(\bphi, t) {\dd}{\mathbf{\bar w}}
\end{equation}
where $\mathbf{\bar w}$ is the inverse Wiener process.

The probability density $p(\bphi,t)$ corresponding to the SDE \eqref{eqn:generalSDE} solves the following Fokker-Planck equation (or Kolgomorov's forward equation) \citep{10.5555/129416}
\begin{equation}\label{eqn:generalFP}
    {\frac{\partial}{\partial t} }p(\bphi(t))
    = -\sum_{i=1}^d  {\partial\over \partial \phi_i } \left( f_i(\bphi,t) p(\bphi(t))\right) +{1\over 2} \sum_{i=1}^d\sum_{j=1}^d {\partial^2\over \partial \phi_i\partial \phi_j }\left(  \sum_{k=1}^d G_{ik}G_{jk}\, p(\bphi(t))\right),
\end{equation}
where $\phi_i$ denotes the component of $\bphi$ in a given basis.
\subsection{Renormalization Group (RG) Flows}
\label{sec:rg}

As mentioned in the introduction, the renormalization group refers to a collection of methods in physics that aim to progressively remove the high-frequency degrees of freedom while retaining the relevant low-frequency  ones. In other words, one aims to remove the irrelevant details of the physical system
without altering the physics at the larger scale one is interested in. By doing so, one hopes to be able to robustly calculate the universal macroscopic features of the physical systems.

There are many ways physicists have proposed to achieve this goal, starting with the seminal work of \citet{PhysicsPhysiqueFizika.2.263} and \citet{PhysRevB.4.3174,PhysRevB.4.3184}.
How to improve the understanding and the implementation of RG flows, including efforts involving machine learning methods, remains an active topic of investigation in physics \citep{koch-janusz2018MutualInformation}.
Here we consider the exact RG (ERG) formalism,  a non-perturbative method pioneered by \citet{Polchinski:1983gv} for quantum field theories. In this RG method, one implements Wilson's idea of RG by specifying a {\it cutoff kernel}
\(K_k(\Lambda):= K(k^2/\Lambda^2)\) for a given \emph{cutoff scale} for each frequency $k$, with the property that $K_k(\Lambda) \rightarrow 1$ when $k\ll  \Lambda$ and  $K_k(\Lambda) \rightarrow 0$ when $k \gg \Lambda$.  With this, one erases information on frequencies much larger than $\Lambda$.
One example of such cutoff kernels is the sigmoid function.

Given a physical theory and a choice of cutoff kernel, one can define physical probability distribution $p_\Lambda[\bphi]$ that satisfies a differential equation which is an infinite-dimensional version of the Fokker-Planck \eqref{eqn:generalFP}, where the role of diffusion time is played by $t=-\log (\Lambda/\Lambda_0)$ for some reference scale $\Lambda_0$.

\subsection{Standard Diffusion Models}

In diffusion-based generative models, a forward diffusion process that gradually transforms data samples into noise following a particularly simple SDE is inverted to transform noise into images.
A commonly used forward SDE is the finite-variance\footnote{This is referred to as the ``variance-preserving" (VP) diffusion in some literature. We will reserve the term to cases when the variance is actually strictly constant throughout the diffusion process, which we will discuss in \S\ref{subsec:whitening}. } SDE \citep{song2021ScoreBasedGenerativeModeling} defined as:
\begin{equation} \label{eqn:SDE_standard1}
    \mathrm{d}\bphi = -\tfrac{1}{2}\beta(t)\bphi\,\mathrm{d}t + \sqrt{\beta(t)}\,\mathrm{d}\mathbf{w},
\end{equation}
where the initial vector $\bphi(0) \in \mathbb{R}^d$ represents the data sample and $\mathrm{d}\mathbf{w}$ denotes the standard Wiener process. The function $\beta: [0,T]\to \mathbb R_{+}$ which determines the SDE is the predefined noise schedule.

The reverse-time SDE follows from specializing \eqref{eqn:generalinverseSDE} and reads
\begin{equation}
    \mathrm{d}\bphi = \left[-\tfrac{1}{2}\beta(t)\bphi - \beta(t)\nabla_\bphi \log p_t(\bphi)\right]\mathrm{d}t + \sqrt{\beta(t)}\mathrm{d}\bar{\mathbf{w}} \,,
\end{equation}
where $\mathrm{d}\bar{\mathbf{w}}$ is a reverse-time Wiener process, and $\nabla_\bphi \log p_t(\bphi)$ is the score function of the marginal distribution at time $t$.
The task for machine learning is thus to approximate the score function, which can be achieved by denoising score matching \citep{vincent2011ConnectionScore} with the objective function
\begin{equation}\label{eqn:loss1}
    \mathcal{L}_{\rm DSM} = \mathbb{E}_{t, \bphi(0), \epsilon} \left[
        \lambda(t) \norm{ s_\theta(\bphi(t), t) - \nabla_{\bphi(t)} \log p_t(\bphi(t) | \bphi(0)) }^2
    \right] \,,
\end{equation}
where  $\epsilon \sim \mathcal{N}(0, \mathbf{I})$ is Gaussian white noise, $\bphi(t) = \alpha(t) \bphi(0) + \sigma(t) \epsilon$ is the noised data at time $t$, $\lambda: [0,T]\to {\mathbb R}_{+}$ is a weighting function, and $\alpha(t), \sigma(t)$, and $\beta(t)$ are functions capturing the equivalent information about the noising schedule. They are defined in \eqref{eqn:parameters1} by specializing $\beta_i=\beta$ etc.
Importantly, the choice of SDE (\ref{eqn:SDE_standard1}) leads to an Ornstein-Uhlenbeck (OU) process and the conditional score $\nabla_{\bphi(t)} \log p_t(\bphi(t) | \bphi(0))$ can be computed analytically \citep{song2021ScoreBasedGenerativeModeling}.

\section{Methods}

\subsection{Diagonalizable Ornstein Uhlenbeck Process
}
\label{subsec:Diagonalizable Ornstein Uhlenbeck Process}
Returning to the general diffusion SDE (\ref{eqn:generalSDE}),
we now consider  the special case in which  $\mathbf{f} ={\boldsymbol F}\bphi$ and ${\boldsymbol F}$ is $\bphi$-independent . This guarantees that the SDE describes a Ornstein-Uhlenbeck process admitting analytical solutions for the conditional distribution $p_t(\bphi(t)\lvert \bphi(0))$ required for denoising score matching.
Moreover, we  consider a choice of  {\it simultaneously  diagonalizable} $\mathbf{F}$ and $\mathbf{G}$:
\begin{equation}\label{eqn:diagonazability}
\boldsymbol{F} =
 M^{-1} {\tilde {{\boldsymbol F}}}   M
,~~  \boldsymbol{G} = M^{-1} \sqrt{\boldsymbol\beta}
\end{equation}
with some constant matrix $M$ and  diagonal $\boldsymbol\beta={\rm{diag}}(\beta_i)$ and $\tilde{\boldsymbol F}={\rm{diag}}(\tilde F_i)$.

In terms of the parameterization
\begin{equation} \label{dfn:chi}
   \boldsymbol \chi := M\bphi ,
\end{equation}
 the SDE \eqref{eqn:generalinverseSDE} is equivalent to $d$ decoupled SDEs of the form
\begin{equation}\label{eqn:chi_diff_general}
    \dd\chi_i =  \tilde{ F}_{i}(t) \chi_i \dd t + \sqrt{{\beta}_i(t)}    \dd{\mathrm w}
\end{equation}
with the reverse SDE given by
\begin{equation}\label{eqn:chi_reverse}
   \dd\chi_i = \left( \tilde{ F}_{i}(t) \chi_i  -{ {\beta}_i(t)}\nabla_{\chi_i} \log p_t(\chi)
   \right) \dd t + \sqrt{{\beta}_i(t)}    {\dd}{\mathrm{\bar w}} \,.
\end{equation}
The choice of the transformation matrix $M$ is a choice of data representation in which the diagonal score-based diffusion based on the SDE \eqref{eqn:chi_diff_general} and \eqref{eqn:chi_reverse} can be efficiently performed.

In particular, as the Wiener process is invariant under orthogonal transformations, it is convenient to view the change of basis (given by $M$) as the composition of an orthogonal ($U$) and a scaling ($S$)  transformation: $M=S^{-1}U$.
In terms of the original $\bphi$ variables, the forward SDE then reads
\begin{equation}
 \dd{\bphi}
  =F\bphi \,dt  +U^{-1} S\sqrt{\boldsymbol \beta} \,{\dd}{\mathrm{ w}}
 =U^{-1}\tilde{\boldsymbol{F}}U\bphi \,dt + \sqrt{\boldsymbol \beta'} \,{\dd}{\mathrm{ w'}}
\end{equation}
where $\sqrt{\boldsymbol \beta'} = U^{-1}\sqrt{\boldsymbol \beta}U$
and $\dd{\mathrm{w'}} = \sqrt{\Sigma_{\mathrm{prior}} } \dd{ \mathrm{w}}$ is a Wiener process with covariance matrix $\Sigma_{\mathrm{prior}} = U^{-1} S^{2}U$.

The choice of $M$, particularly the orthogonal part $U$, captures the freedom in our unified framework to choose the {\bf basis} in which the diffusion process is diagonal.
Moreover, the scaling $S$ then determines the choice of the noise ({\bf prior}) distribution $p_{\mathrm{prior}}=\mathcal{N}(0, \Sigma_{\mathrm{prior}})$, which the forward process approaches at late times. Finally, note that the $\beta_i(t)$ can a priori be independent functions of $t$ for each component $i$. The choice of $\beta_i(t)$ thus captures the choice of a {\bf component-wise noising schedule}.

Due to the diagonal property of the SDE, the denoising score matching loss function \eqref{eqn:loss1} for learning the Stein score  $\nabla_{\chi_i} \log p_t(\chi)$ can be straightforwardly generalized to the GUD models:
\begin{equation}\label{eqn:lossGUD}
    \mathcal{L}_{\rm GUD} = \mathbb{E}_{t, \bchi(0), \epsilon} \sum_{i=1,\dots,d}
        \lambda_i(t) \Big\lvert s_{i,\theta}(\bchi(t), t) - \nabla_{\chi_i(t)} \log p_t(\bchi(t) | \bchi(0)) \Big\lvert^2
     \,
\end{equation}
where the $\boldsymbol \lambda = (\lambda _1,\dots,\lambda_d ) : [0,T]\to {\mathbb R}_{+}^d$ is the weighting vector.
In our experiments, we let
$\lambda_i(t) = \sigma_i^2(t)$,
with the aim to scale the loss to be an order-one quantity and
generalizing
 the common weighting factor $\lambda(t)=\sigma^2(t)$ in the standard diffusion loss (\ref{eqn:loss1}) \citep{song2020GenerativeModelingEstimating}.
The SDE (\ref{eqn:chi_reverse}) with the learned score, when discretized, leads to a hierarchical generative model with model density
\begin{equation}\label{eqn:discrete}
    \tilde p(\bchi(0))= \int \left(\prod_{k=1}^T d^d\bchi(\tfrac{k}{T}) \right)\;
    \left(\prod_{\ell=0}^{T-1} p(\bchi({\tfrac{\ell}{T}})\lvert \bchi({\tfrac{\ell+1}{T}}))\right)
  \tilde p(\bchi(1))
\end{equation}
where $T$ is the number of steps in the discretization, and the prior distribution is given by the noise distribution
\( \tilde p(\bchi(1)) = {\cal N}(0,{\bf I})
\).

\subsection{Finite-Variance Diffusion and the Signal to Noise Ratio}
\label{sec:noising-variables}

In the coordinate given by $\bchi$, we now further specialize \eqref{eqn:chi_diff_general} to the following  finite-variance diffusion process:
with  \(\tilde {\boldsymbol{F}}(t) = -{1\over 2} \bbe(t)\), the corresponding SDE reads
\begin{equation}\label{eqn:chi_diff}
    d\chi_i =  -{1\over 2} {\beta}_i(t) \chi_i dt + \sqrt{{\beta}_i(t)}    d{\mathrm w}.
\end{equation}
Integrating the above gives
\begin{equation}
 \label{SDE_integrated}
  \chi_i(t) =    \alpha_{i}(t)  \chi_i(0) + \sigma_i(t)  \epsilon \;,\qquad \epsilon\sim {\cal N}(0,1)
 \end{equation}
where
\begin{equation}\label{eqn:parameters1}
    \alpha_{i}(t) = \exp(-\frac{1}{2} \int_0^t \beta_i(s)\dd{s})\,, \quad \text{and} \quad
   \sigma_i(t)^2 = 1-\alpha_i(t)^2
    \,.
\end{equation}
It follows that the variance
\begin{equation}
{\rm Var}(\chi_i(t)) = \alpha_i(t)^2(\Sigma^{(\bchi)}(0))_{ii} + \sigma_i(t)^2
\end{equation}
interpolates between 1 and the data variance $(\Sigma^{(\bchi)}(0))_{ii}$, and is in particular finite at all stages of diffusion.
In the above, we have used the following notation for the data covariance matrix
$$(\Sigma^{(\bchi)}(0))_{ij} := {\mathbb E}_{p_{\rm data}(\bchi(0))} \left[ (\chi_{i}(0) -\overline{\chi_{i}(0)} )
    (\chi_{j}(0) -\overline{\chi_{j}(0)} )\right]
    ~,~ {\rm where}\quad \overline{\chi_{i}(0)}:=  {\mathbb E}_{p_{\rm data}(\bchi_0)} [\chi_{0,i}],
    $$

An important quantity signifying the stage of the diffusion process (for each component) is the time evolution of the ratio between the signal and the noise, captured by the {\emph{signal-to-noise ratio}},
\begin{equation}\label{dfn:SNR1}
    {\rm SNR}_i(t):= {\mathbb E}\left( \frac{ (\alpha_{i}(t)  \chi_i(0))^2 }{\sigma_i(t)^2}\right)
=(\Sigma^{(\bchi)}(0))_{ii}  \frac{\alpha_i^2(t)}{\sigma_i^2(t)} ,
\end{equation}
where the expectation is with respect to the data and the noise distribution, and we have assumed that the data mean vanishes (which can always be made to be the case by subtracting the mean).
Note that this is different from the signal-to-noise ratio quoted in some diffusion model contexts,
\begin{equation}\label{eqn:gamma}
    {  {\rm snr}}_i(t) :=   {\alpha_i^2(t)}/{\sigma_i^2(t)}= e^{-\gamma_i(t)},
\end{equation}
as this version does not take into account the magnitude of the signal in the data.
As they depend only on the schedule, we note that the functions
$\beta_i$,  $\alpha_i$, $\sigma_i$ and $\gamma_i$ all contain the same information.

\begin{wrapfigure}[12]{r}{0.3\textwidth}
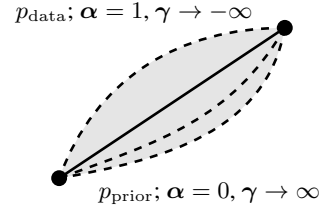

    \vspace{-9pt}
    \include{distribution-path}
\vspace{-30pt}
\caption{\label{fig:schematic_schedule}\footnotesize Different noising schedules
$\boldsymbol \gamma(t)$.}
\end{wrapfigure}

At a given time $t\in [0,T]$ in the diffusion process, the information of ${\boldsymbol\gamma}(t)=(\gamma_1,\dots,\gamma_d)(t)\in {\mathbb R}^d$ is what we call the {\emph{noising state}}, indicating the extent to which information in the data has been replaced by noise at that time. As a function of $t$, the evolution of the noising state traces out a path connecting the data distribution $p_{\rm data}$, corresponding to $${\boldsymbol \alpha}({\boldsymbol\gamma}) = (\alpha_1,\dots , \alpha_d) =(1,\dots,1) ,$$
and the prior distribution  $p_{\rm prior}$, corresponding to ${\boldsymbol \alpha} =(0,\dots,0)$, where $\alpha_i={\rm sigmoid}(-\gamma_i)^{1/2}$ as in \eqref{eqn:gamma}.
The different paths correspond to different
Ornstein-Uhlenbeck processes, as defined in  \eqref{eqn:chi_diff}, with different diffusion dynamics.

This highlights the fact that the freedom in component-wise noising schedules in the GUD model is fundamentally larger than the
freedom in the noising schedule in standard diffusion models, which is given by different choices of the function $\gamma(t)$ with $\gamma_i(t)=\gamma_j(t)=\gamma(t)$.
In this case, all choices of $\gamma(t)$ trace out the same diagonal path (as long as the boundary values $\gamma(0)$, $\gamma(T)$ are held fixed) and merely amount to different time parameterizations \citep{kingma2023VariationalDiffusion}. In contrast,  different component-wise schedules generically correspond to genuinely different paths, as illustrated in the schematic Figure \ref{fig:schematic_schedule} where the continuous line corresponds to the standard diffusion schedule (with any time parametrization) and the dashed lines represent other possible component-wise schedules.

\subsection{Unification via Soft-Conditioning}
The above form of SNR clarifies an implicit hierarchical structure of the standard diffusion models: even when $\gamma_i(t)=\gamma(t)$ is identical for all components $i$, the components with larger amplitudes have larger signal-to-noise ratio {SNR$_i(t) =(\Sigma^{(\bchi)}(0))_{ii}  e^{- \gamma(t)}$, and are in this sense less ``noised" throughout the diffusion process. As a result, the generation process \eqref{eqn:discrete} and in particular the modeling of the probability $p(\bchi({\tfrac{\ell}{T}})\lvert \bchi({\tfrac{\ell+1}{T}}))$ conditional on the previous state is implicitly a process of generating the less important features (with smaller amplitude) conditional on the more important features (with larger amplitude) that have already been partially generated. It is clear that by making more general choices of component-dependent noising schedules $\gamma_i(t)$ one can tune the degree of this {\emph soft-conditioning} property, as we will explore in the experiments below. In the extreme case when the support of $\beta_i(t)$ and $\beta_{i\neq j}(t)$, namely the ``active time"  for the  $i$th resp. $j$th component, do not overlap, we arrive at autoregressive generation, in which one feature/token (or one group of features/tokens) is generated at each time, conditional on those that have been generated already. See  Figure \ref{fig:column-noising} for the visualization of a specific example.
In this way, the freedom to choose a component-dependent noising schedule in our GUD model enables us to interpolate between standard diffusion and autoregressive generation.

\subsection{Whitening}\label{subsec:whitening}
A particularly interesting choice for the matrix $M=S^{-1}U$ is the orthogonal transformation $U$ that diagonalizes the data  covariance matrix $\Sigma^{(\bphi)}(0)$, and the diagonal matrix $S^{-1}$ that performs a {\emph{whitening}} transformation.
In other words, we choose $S$ and $U$ such that the data covariance matrix matches $\Sigma^{(\bphi)}(0) = U^{-1} S^2 U$.
Note that $M$ is then precisely the familiar PCA transformation followed by a whitening transformation which makes the variance uniform.
In the context of diffusive generation, such a basis has the following appealing features. First, the softness of the soft-conditioning, manifested via the evolution of the signal-to-noise ratio \eqref{dfn:SNR1}, is now completely controlled by the component-wise schedule $\gamma_i(t)$, which can make the design process of the diffusion modeled more streamlined and uniform across different applications with different datasets. Second, with such a choice the covariance matrix actually remains constant throughout the diffusion process as the data covariance $ \Sigma^{(\bchi)}(0) =  {\bf I} $ is now the same as the noise covariance, and the finite-diffusion \eqref{eqn:chi_diff} is {\emph{variance preserving}} in the strict sense. In other words, the conditional number of the covariance matrix is always one and the generative process does not need to alter the second-order statistics. We expect this property to be beneficial in some situations for learning and discretization.

\subsection{Noising-State Conditional Network Architecture}

For the score network architecture, we follow the approach of predicting the noise $\epsilon$ given a noised image \citep{ho2020DenoisingDiffusionProbabilistic},
trained via denoising score matching \citep{vincent2011ConnectionScore}.
In standard diffusion models, this score network is typically conditioned on the time variable or an equivalent object such as $\gamma(t)$ \citep{kingma2023VariationalDiffusion}.
The introduction of a component-wise schedule in our framework suggests generalizing this by conditioning the model on the more informative component-wise noising state, represented by the component-wise noise state ${\boldsymbol\gamma}(t)=(\gamma_1,\dots,\gamma_d)(t)$.
Since this is a vector of the same dimension as the data and not a scalar, a modification of the network architecture is required.
We have implemented this by incorporating cross-attention between the data and the noising state, further details can be found in section \ref{sec:architecture} of the appendix.

Since any choice of the noising schedule $\gamma_i(t)=\gamma(t)$ in standard diffusion models can be thought of as just a reparametrization of time \citep{kingma2023VariationalDiffusion} (cf. \S\ref{sec:noising-variables}), the diffusion time $t$ itself suffices as a feature for the network to indicate the noising state.  For our GUD models, this is true only for a fixed schedule choice.
By conditioning directly on $\boldsymbol\gamma$ instead of $t$, our score network is directly conditioned on the instantaneous noising state, and not on the totality of its path, namely the schedule ${\boldsymbol\gamma}(t)$.
This enables us to train a single network for a range of schedules, as we will do in the experiments described in the next section.
The set of values ${\boldsymbol\gamma}\in {\mathbb R}^d$ used during training bound a region, visualized schematically by the shaded area in Fig. \ref{fig:schematic_schedule}.
This is implicitly the region of the values of $\boldsymbol \gamma \in \mathbb{R}^d$ where the score function has been learned. This suggests the possibility of using any particular path within the shaded region for generation, which might differ from the path used for training (indicated by the dashed lines in Fig. \ref{fig:schematic_schedule}).
This feature of the GUD model may facilitate the numerical optimization of component-wise schedules in future work.

\section{Experiments}

We will now showcase the flexibility of the GUD model with some examples, and conduct preliminary investigations into the effects of these different design choices on the behavior of diffusion models and their resulting sampling quality.
An overview of the experiments, highlighting the relevant design choices, is given in the following table.
\begin{table}[H]
\centering
\resizebox{.9\columnwidth}{!}{%
\begin{tabular}{c|c|c|c}
& \S \ref{sec:autoregressiveness} & \S \ref{sec:column} &\S \ref{sec:haar}\\\midrule\vspace{2pt}
basis & pixel, PCA , FFT  & column & wavelet $\otimes$ column \\  \midrule
\multirow{2}*{prior} & isotropic Gaussian and &\multirow{2}*{isotropic Gaussian } &\multirow{2}*{isotropic Gaussian } \\
& variance-matching Gaussian&&\\
\midrule
\multirow{2}*{noising schedule} & varying softness  &\multirow{2}*{varying softness} &\multirow{2}*{varying softness} \\
& and ordering variables && \\
\midrule
other applications &  & image extension& \\
\end{tabular}}
\end{table}

\subsection{Soft-Conditioning Schedules}
\label{sec:autoregressiveness}

\begin{figure}[tbh]
    \centering
    \includegraphics[width=.85\linewidth]{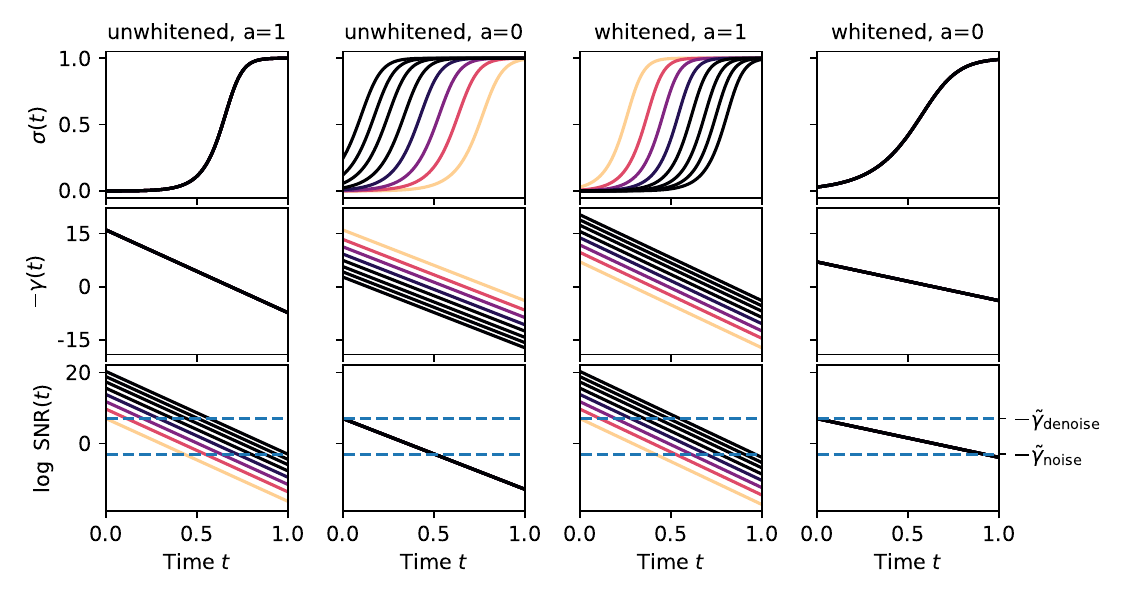}
    \vspace{-15pt}
    \caption{\footnotesize
    For eight of the PCA components $\chi_i$ of CIFAR-10, we
    visualize the OU noise level ${\sigma}_i(t)$, the corresponding noising path ${\gamma}_i(t)={\rm logit}(\sigma_i^2(t))$ for the linear schedule of \eqref{eq:linear-schedule}, and the corresponding signal-to-noise ratio. Blue dashed lines indicate chosen minimal noising/reconstruction levels. From left to right: (a)  Standard diffusion where ${\gamma}_i={\gamma}_j$. (b) The schedule $\boldsymbol \gamma$ is chosen such that $\log{\rm SNR}_i(t)=\log{\rm SNR}_j(t)$, corresponding to a generative process with no hierarchy. (c) With whitened data, with the schedule $\boldsymbol \gamma$ chosen such that $\log{\rm SNR}_i(t)$ is identical to that in the standard diffusion case shown in column (a). (d) Hierarchy-less generation with whitened data. }
    \label{fig:schedules-overview}
\end{figure}

First, we investigate the effect of choosing different bases, priors, and schedules by conducting experiments on the unconditional generation of CIFAR-10 images.
As an example of a  simple setup, we choose linear noise schedules, which means that all $\gamma_i(t)$ we consider here are linear functions of the diffusion time $t$.
In the first setup, we choose the basis, given by the orthogonal transformation $U$ described in  \S\ref{subsec:Diagonalizable Ornstein Uhlenbeck Process},  to be the PCA basis.
In terms of prior distributions, we choose our noise to be either given by isotropic Gaussian or Gaussian with covariance matching that of the data. Alternatively, as explained in \S\ref{subsec:Diagonalizable Ornstein Uhlenbeck Process}, this is equivalent to whitening the data (in the PCA basis) while using the isotropic Gaussian noise. We will therefore refer to the two choices of priors as {\emph{ whitened}} and  {\emph{ unwhitened}} in what follows.

In the second setup, we use the Fourier basis and consider a two-parameter family of component-wise noising schedules, where we vary the precise ordering of the different Fourier components (given by the ``ordering variables") as well as the softness parameter of the soft-conditioning schedule.
It is important to note that these experiments only constitute a one- or two-dimensional subspace of a much larger design space. We make the above choice for the sake of concreteness and due to a limitation in computational resources.

\paragraph{Linear component-wise schedules.}
For a given choice of basis $\bchi$, let $\Sigma_i:=(\Sigma^{(\bchi)}(0))_{ii}$ denote the variance of each component $\chi_i$. As discussed in  \S\ref{sec:noising-variables}, a choice of the schedule $\boldsymbol\gamma(t)$ then leads to the signal-to-noise ratio  $ \log \mathrm{SNR}_i(t) = -\gamma_i(t) + \log \Sigma_i$.
For the diffusion generative models, each component must be sufficiently noised in the forward process and sufficiently denoised in the reverse process.
It will therefore be convenient to introduce the parameters $\tilde{\gamma}_{\mathrm{denoise}}$ and $\tilde{\gamma}_{\mathrm{noise}}$
which specify the minimal levels of denoising and noising at the initial and final time, respectively:
\begin{gather}
\notag
\begin{split}
\tilde{\gamma}_{\mathrm{denoise}} &= -\min\limits_{i}   \log \mathrm{SNR}_i(t=0) \\\tilde{\gamma}_{\mathrm{noise}} &= -\max\limits_i \log \mathrm{SNR}_i(t=1) .
\end{split}
\end{gather}
Here and in what follows we parametrize the diffusion time to be $t\in[0, T]=[0, 1]$.
The parameters $\tilde{\gamma}_{\mathrm{denoise}}$ and $\tilde{\gamma}_{\mathrm{noise}}$ have to be chosen carefully.
In particular, to employ the inverse process as a generative model, the distribution at the final time $t=1$ must be sufficiently close to the prior normal distribution from which we draw initial samples. This is not guaranteed by demanding endpoints for the SNR alone, as a low SNR can be achieved with little noise if the data magnitude is small. We therefore also require $\sigma_i(t=1) \geq \sigma_{\min}$, which translates to $\tilde{\gamma}_{\mathrm{noise}} \geq {\rm logit}(\sigma_{\min}^2)- \min_i\log\Sigma_i$.

Next, we associate an ordering variable $l_i$ to each component, which will determine the hierarchical structure among the components by shifting the onset of noise in the forward process. CIFAR-10, just like many other (natural) image datasets, is an example of what might be called
frequency-based datasets. We use this term to describe the datasets with a natural meaning of locality, whose covariance is approximately diagonalized in the Fourier basis and whose variance is generally decreasing with increasing frequencies \citep{Tolhurst,Field87}. For these datasets, the hierarchical structure can naturally be specified in terms of the related notions of variance, frequencies, and resolution, as familiar from image processing.
We choose our ordering variable $l_i$ to capture this notion of hierarchy.

To control the level of autoregressiveness, or softness, of the soft-conditioning linear schedule, we introduce a parameter $a > 0$ and define the linearly interpolating schedule
\begin{equation}
    \label{eq:linear-schedule}
    \gamma_i(t) = \gamma_{\min,i} +  (\gamma_{\max,i} - \gamma_{\min,i}) t
\end{equation}
with the endpoints given by
\begin{equation}
\begin{gathered}
    \gamma_{\min,i} := \tilde{\gamma}_{\mathrm{denoise}}+\log\Sigma_i+a(l_i - l_{\max}) \, \\
    \gamma_{\max,i} := \tilde{\gamma}_{\mathrm{noise}}+\log\Sigma_i+a(l_i - l_{\min})\,.
\end{gathered}
\end{equation}
\begin{wrapfigure}[18]{r}{0.47\textwidth}
    \centering
    \vspace{-11pt}
    \includegraphics[width=\linewidth]{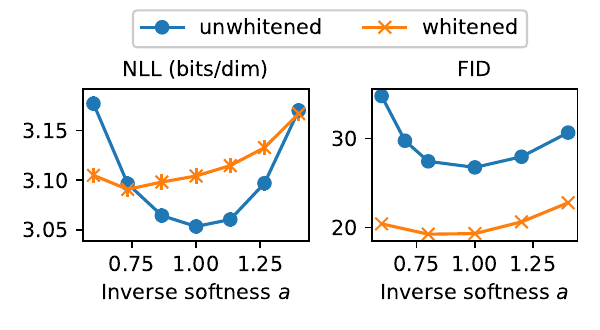}
    \vspace{-15pt}
    \caption{\footnotesize Dependence of model quality in terms of negative log-likelihood (left) and FID (right) on the softness parameter for the linear schedule in \S \ref{sec:autoregressiveness}. The schedule is defined in PCA components and results are shown both for unwhitened and whitened data scaling (i.e.~white and data-matching priors). Training on CIFAR-10 using a single score-network for each choice of scaling. Standard diffusion corresponds to $a=1$ in the unwhitened case.}
    \label{fig:cifar10-scan-pca-s}
\end{wrapfigure}
Here $l_{\rm max} = \max_i l_i$ and similarly for $ l_{\rm min}$.
The larger $a$ is, i.e.~the smaller the softness $1/a$, the more autoregressive the schedule becomes. On the other hand, in the limit of extreme softness (small $a$) the hierarchical nature of the generative model disappears.

The parameters of our linear schedules are thus the ordering variables $l_i$, the softness parameter  $a^{-1}$, and the SNR endpoints given by $\tilde{\gamma}_{\rm denoise}$, $\tilde{\gamma}_{\rm noise}$.
In our experiments, we fix $\tilde{\gamma}_{\rm denoise}=-7$ and $\tilde{\gamma}_{\rm noise}=\max[3,{\rm logit}(\sigma_{\min}^2)- \min_i\log\Sigma_i]$ with $\sigma_{\min} = 0.99$.

\paragraph{Softness in PCA space.}
In the first experiment,
we apply the above linear schedule in the whitened and unwhitened PCA bases.
We choose the ordering variable to be given by $l_i = -  \log \Sigma_i$, which allows for the trajectory of the signal-to-noise ratio of the standard diffusion model to be reproduced at $a=1$, also when the prior has been changed to have the same covariance as the data (see Fig. \ref{fig:schedules-overview} (c)).
We trained a single score network for the (inverse) softness parameter in the range $a \in [0.4, 1.6]$ by randomly sampling $a$ at each training step.
Figure \ref{fig:cifar10-scan-pca-s} shows the negative log-likelihood (NLL) and FID evaluated for different values of $a$\footnote{The FID has not converged and was still dropping at a significant rate when we terminated training due to limited computational resources. Nevertheless, the qualitative features displayed in Figure \ref{fig:cifar10-scan-pca-s} appear to have stabilized. }. Interestingly, the unwhitened configuration performs better when measured by NLL, but worse in terms of FID, with the standard diffusion setup ($a = 1$) appearing close to optimal.
See section \ref{sec:training} of the appendix for further experimental details.

\paragraph{Ordering variables.}
We also perform experiments in the Fourier (FFT) basis, for which the RG physics reviewed in \S\ref{sec:rg} naturally suggests an ordering of noising based on the frequency $|k_i|$.
To test the dependence of the quality of the model on ordering parameters, we consider ordering variables
$l_i = (1 - r) (- \log \Sigma_i) + r (|k_i| + \delta)/\kappa $, parametrized by $r\in[0,1]$ and interpolating between   $l_i=- \log \Sigma_i$ as in the PCA experiments and the frequencies $|k_i|$. The slope and offset parameters, $\kappa$ and  $\delta$,  are chosen such that the range of $l_i$ are the same at $r=1$ as at $r=0$. We trained a score network for a range of values of $r$ in addition to the softness parameter $a^{-1}$ on CIFAR-10, with evaluation results shown in Figure \ref{fig:cifar10-scan-fft-rs}. We find the optimal performance in terms of NLL is located slightly away but close to standard diffusion. In this experiment, we choose the prior to be the isotropic Gaussian (i.e.~unwhitened).

\subsection{Sequential Generation in Real Space}
\label{sec:column}

\begin{figure}[tb]
    \centering
\vspace{-32pt}
\includegraphics[width=1\linewidth]{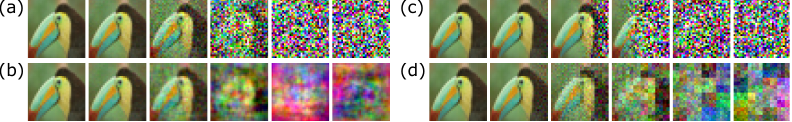}

\vspace{-8pt}
    \caption{\footnotesize Diffusion forward process for a single image of CIFAR-10: (a) standard diffusion, (b) variance-matching Gaussian noise with same SNR as standard diffusion, (c) column-wise sequential schedule of \S \ref{sec:column} with $b=0.5$, (d) combination of Haar wavelet and column-sequential schedule of \S \ref{sec:haar} with $a=0.5$, and with variance-matching Gaussian noise.}
    \label{fig:fwd}
\end{figure}

\begin{figure}[tb]
    \centering
    \begin{minipage}[t]{0.5\linewidth}
        \centering
\includegraphics[width=\linewidth]{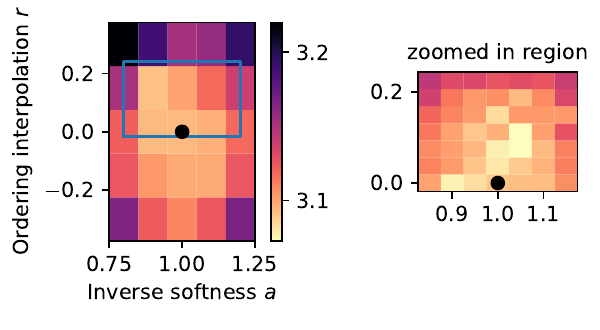}

\vspace{-15pt}
        \caption{\footnotesize The model quality with a two-parameter family of schedules controlling the softness and the ordering parameters.
 The right figure is the region in the box on the left and the same color map is shared. The black dots indicate the parameters corresponding to standard diffusion models.}
        \label{fig:cifar10-scan-fft-rs}
    \end{minipage}
    \hfill
    \begin{minipage}[t]{0.47\linewidth}
\centering\includegraphics[width=\linewidth]{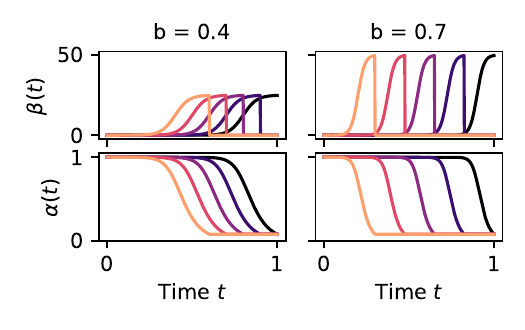}

\vspace{-15pt}
        \caption{\footnotesize An illustration of column-wise schedule of  \S\ref{sec:column} for $5$ columns and different softness parameter $b^{-1}$. The larger $b$, the more autoregressive the model is, as the overlap of the ``active'' times with  noising rate $\beta_i>0$ decreases, and similarly for the suppression factors $\alpha_i$.
}
        \label{fig:column-noising}
    \end{minipage}
\end{figure}

While the previous experiment explores the GUD model in the context of multi-scale hierarchical generation, it can equally be applied to perform sequential generation in pixel space, as we will now demonstrate with a soft-conditioning column-wise generation model.
Grouping the components according to their column in the pixel space of size $L\times L$, we index the schedules according to the columns labeled by $i=1, \ldots L$.

As before, we perform experiments with a linear schedule
 \begin{equation}
\label{eq:column-schedule}
   \gamma_i(t) = \mathrm{clip}_{\gamma_{\min}, \gamma_{\max}}\qty(\gamma_{\min} +(t- t_i) \frac{\gamma_{\max} - \gamma_{\min}}{1 - b})
~,~~~{\rm with}~~t_i =  b \frac{L - i}{L-1}\end{equation}
 with a varying degree of softness/autoregressivity, now parametrized by $b$.
The clipping, defined by the clipping function  ${\rm clip}_{y,z}(x) := \max(y, \min(z, x))$, has the effect of freezing the columns when the designated noising ($\gamma_{\rm max}$) or reconstruction level ($\gamma_{\rm min}$) is reached.

\paragraph{Training on PCAM dataset.}
We trained separate score networks at $b=0.3$ and $b=0.5$ on the PCAM dataset \citep{Veeling2018-qh}, downscaled to $32 \times 32$ pixels, obtaining negative log-likelihoods of $3.90$ and $3.94\,\mathrm{bits/dim}$, respectively.

\paragraph{Image extension.}
Since images are generated column-by-column from left to right, conditioned on the part on the left that has already been generated, this suggests that the generative process can be repeated. See section \ref{sec:column-detail} of the appendix for further detail.
Figure \ref{fig:pcam-texture-gen} shows three examples of image strips generated in this manner, using the score network trained on $32 \times 32$ images at $b = 0.5$.

\begin{figure}[ht]
    \centering
    \vspace{-10pt}
\includegraphics[width=.85\linewidth]{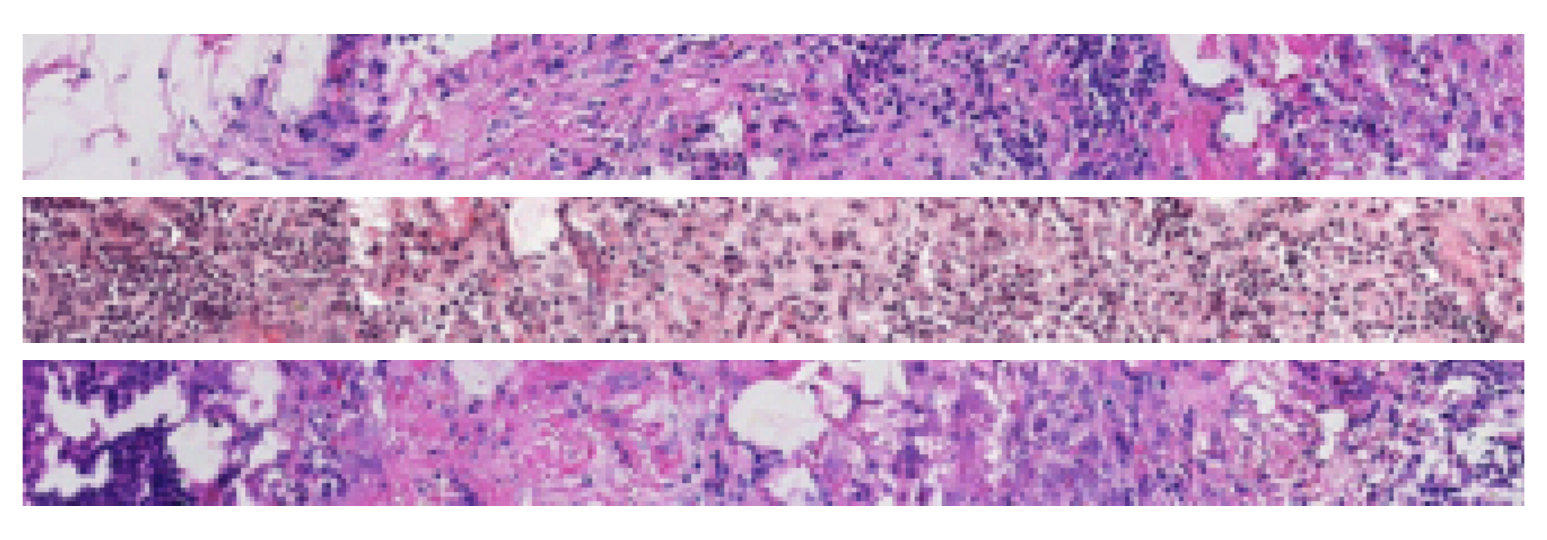}

    \vspace{-15pt}
    \caption{ \footnotesize Images generated using the column-wise schedule by repeatedly adding noise to the right, using a score network that was only trained on square training data.}
    \label{fig:pcam-texture-gen}

     \vspace{-15pt}
\end{figure}

\subsection{Haar wavelets}
\label{sec:haar}

To further showcase the versatility of our unified framework, we integrate Haar wavelet decomposition with a column-wise noise schedule among the wavelet components at each hierarchical level. This extends the wavelet-conditioned score matching of  \citet{guth2022WaveletScoreBased} by including a parameter allowing for soft-conditioning, and incorporating column-wise sequential noising at each level.

Concretely, we use two parameters $a$ and $b$ to parametrize the (inverse) softness among the different levels of wavelet components and the columns within each level, respectively.
Suppose there are $\mathcal{N}$ hierarchical levels of wavelet decompositions, labeled by $i=1,\ldots, \mathcal{N}$, and there are $L_i$ columns in the $i$-th level, indexed by $j=1,\ldots,L_i$, we define the offsets \[
c_i = a \frac{\mathcal{N} - i}{\mathcal{N} - 1}, ~~ c_{ij} = b \frac{L_i - j}{L_i - 1}.
\]
With this, we specify the linear schedule for $a,b\in[0,1]$ to be
\[
\gamma_{ij}(t) = \mathrm{clip}_{\gamma_{\rm min}, \gamma_{\rm max}}\qty(\gamma_{\rm min} + (\gamma_{\rm max} - \gamma_{\rm min}) \frac{t_i - c_{ij}}{1 - b} ) ,
\]
where
\(
t_i = \mathrm{clip}_{0, 1}(t - \frac{c_i}{ 1 - a}).
\)
See Figure \ref{fig:fwd}(d) for a visualization of this schedule.

We trained a score network for  $a\in [0.3, 0.7]$ and $b\in [0.3, 0.7]$ on CIFAR-10 for $300$k steps and using $\mathcal{N}=3$ levels. Similar to the results in \S\ref{sec:autoregressiveness},
the model quality again depends on the softness parameters, with the lowest NLL value reached being $3.17 \, \mathrm{bits/dim}$.

\section{Conclusions}
In this work, we proposed the GUD framework, which naturally integrates novel design freedoms in diffusion-based generation. Notably, the framework eliminates the rigid boundary between diffusive and autoregressive generation methods and instead offers a continuous interpolation between the two. This flexibility paves the way for a broad range of potential applications.

First, our experiments indicate that the choices in all three aspects we investigate in the present work -- the diagonal basis, the prior distribution, the component-wise schedule -- do have an influence on the final quality of the model. As a result,  there is potentially vast room to improve the quality of diffusion models. In future work, we will address the question of the optimization of these design choices.

Second, the flexibility of our framework enables seamless integration of various approaches to generative models.
For instance, we illustrated in \S\ref{sec:haar} the possibility to combine hierarchical generation (in the wavelet basis) with sequential generation, and in \S\ref{sec:column} how our framework can readily be used to extend images.  Similarly, the inpainting, coloring, upscaling, and conditional generation tasks can all be realized and generalized within the GUD framework, via an appropriate choice of basis and component-wise schedules.

While the scope of our numerical experimentations and our ability to optimize important hyperparameters has been limited by the compute resources available to us, we believe our theoretical framework has the potential to lead to more efficient diffusion models, a wide range of applications, and novel architecture designs.

\section*{Acknowledgements}
This work has been supported by the Vidi grant (number 016.Vidi.189.182) from the Dutch Research Council (NWO).
MG was partially supported by a project that has received funding from the European Research Council (ERC) under the European Union's Horizon 2020 research and innovation programme (Grant agreement No. 864035).

\bibliography{DiffusionRG}
\bibliographystyle{iclr2025_conference}

\appendix

\section{Noising state conditional score network architecture}
\label{sec:architecture}

Inspired by techniques in conditional generation tasks \citep{rombach2022HighResolutionImage}, we introduced a cross-attention mechanism between the intermediate embeddings of the image and the component-wise noising state $\boldsymbol\gamma$, allowing the network to effectively modulate its predictions based on the noising state at each stage of the diffusion process.
Otherwise, we follow a U-Net architecture similar to \citet{song2021ScoreBasedGenerativeModeling},

To incorporate additional information on the structure of the data, we first concatenate the noising state $\boldsymbol\gamma$ with position labels specific to the application and the choice of basis. For instance, for the PCA example in \S \ref{sec:autoregressiveness}  this is the negative logarithm of the variance of each component, which is also used as the ordering variable. For the experiment in Fourier space, we used the FFT frequency label $|k|$.
In \S\ref{sec:column}-\ref{sec:haar} we used a sequence of integers which increments by one for each subsequent column and adjacent group of Haar wavelet components, respectively.
The concatenated inputs are then processed through MLP-Mixer layers to facilitate learned embeddings of $\boldsymbol\gamma$.
Along the depth of the U-Net, a single dense layer is used to reduce the spatial extent to that of the coarse images at that level, before they are input into the cross-attention.

\section{Experimental Details}
\label{sec:training}

\paragraph{PCA and Fourier bases.}
In our experiments we make use of datasets of colored images, which  have pixel and color channel indices.
Among our choices of basis is the Fourier basis.  The Fourier transform (specifically the fast Fourier transform) is applied independently in each color channel. To have an analogous PCA basis, we have decided to perform the same orthogonal transformation -- the one corresponding the PCA basis of the color-averaged data -- in each color channel. It could be interesting to investigate further choices, including the PCA transformation that mixes the color channels.

\paragraph{Training.}
Unless specified otherwise, training was done with a batch size of $128$ using the Adam optimizer with a learning rate of $5\times10^{-4}$. The validation parameters used to evaluate sample quality are exponentially moving averages updated at a rate of $0.999$.
The diffusion times for denoising score matching are sampled uniformly in $[0, 1]$, and schedule parameters (where applicable) were drawn uniformly from the specified range for each training batch of samples.

The score networks for both CIFAR-10 and PCAM were trained for $300$k training steps on NVIDIA A100 and H100s.
After this time the NLL appears to have converged. However, the FID in the case of CIFAR-10 keeps decreasing with further training.
For the experiments of \S \ref{sec:autoregressiveness} training was therefore resumed and continued for another $100$k steps and with a batch size of $512$ samples.
The FID has apparently not yet reached its minimum for the shown values, and we intend to update these once we have gained access to further computational resources.

\paragraph{Dataset processing.}
We have used a uniformly dequantized version of the dataset, both for training and evaluation, by first adding uniform noise to each quantized pixel value and then rescaling it to $[-1, 1]$.
We have additionally removed the empirical mean of the dataset, computed on all training data.
Otherwise, the mean would have to be taken into account when defining the magnitude-sensitive SNR, instead of just the variances as discussed in \S \ref{sec:noising-variables}, and dividing by the mean when ``whitening'' could lead to extremely large values when the variance of a component is much smaller than its mean.

\paragraph{Scores representations.}
The scores in different data representations, e.g. the original data $\bphi$ and the chosen components $\bchi$ in the notation of \S\ref{subsec:Diagonalizable Ornstein Uhlenbeck Process}, are related by
\begin{equation}
    \label{eq:score-equivalence}
    \nabla_{\bchi} \log p_t(\bchi) =   S  U^\dagger \nabla_\bphi \log p_t(\bphi) \,,
\end{equation}
and we can always go back-and-forth between the scores in both basis.

As we base our architecture on the commonly used convolutional U-net architecture  as in \citep{song2021ScoreBasedGenerativeModeling}, which implicitly assumes the locality and approximately shift-symmetric properties, we let the inputs and outputs of the score network to be always represented in the original image space and not the chosen PCA, Fourier, or other basis.

\paragraph{Evaluation and sampling.}
Evaluation of the negative log-likelihood was done using the ODE corresponding to the reverse SDE \eqref{eqn:chi_reverse}.
Specifically, the SDE of \eqref{eqn:chi_diff}, i.e.
\begin{equation}
    d\chi_i =  -{1\over 2} {\beta}_i(t) \chi_i \dd t + \sqrt{{\beta}_i(t)}    \dd{\mathrm w}
\end{equation}
has a corresponding deterministic ODE that produces the same marginal probabilities, given by
\begin{equation}
    d\chi_i =  - \frac{1}{2} {\beta}_i(t) ( \chi_i  + \nabla_{\chi_i} \log p_t(\bchi) ) \dd{t} \,.
\end{equation}
We use the above ODE to compute the log-likelihoods of the data under the trained models, with the score above replaced by its learned approximation (for more details we refer to \citet{song2021ScoreBasedGenerativeModeling}).
We used $6144$ samples in each computation, averaging over $3$ different slices of the Hessian in Hutchinson's trace estimator for each sample.
For the integrator, we used Tsitouras' 5/4 method as implemented in \citet{kidger2021on} with adaptive step size and both relative and absolute tolerance of $1\times 10^{-4}$.

\section{Repeated Column-Wise Generation}
\label{sec:column-detail}

The real-space sequential column of \S \ref{sec:column} generates images conditional on the left part of the image that has already been denoised, after an initial stage in which the first columns get denoised.
This immediately suggests an application in reconstructing an image that is only partially available.
Figure \ref{fig:pcam-reconstruct} shows how partially noised images can be reconstructed by filling in the right hand side of the image.
Different choices of random key then generate slightly different completions.

The linear column-wise schedule in \eqref{eq:column-schedule} is defined such that by integrating the diffusion process by a time $\Delta t = b / (L - 1)$, the ``noising front'' is effectively moved by one pixel. In other words, in the forward process, a particular column after this time has reached the same SNR its left neighbor had at the previous time.
Starting from an image at noising time $t > 0$, we can generate in principal infinitely long strips of texture.
First, we denoise using the learned score from $t$ to $t + k \Delta t$ with $k<L$ a positive integer.
Then, we cut off the first $k$ columns and store them for the final image.
Next, we append $k$ columns of noise drawn from the prior to the right of the image.
As long as the softness parameter $b$ was chosen sufficiently large given the particular $k$, this effectively restores the image to the noising state at the original time $t$.
This process can thus be repeated, and by concatenating the previously generated left-side columns, a connected rectangular stripe of image is constructed.
As an example, in Figure \ref{fig:pcam-texture-gen} we show the results with $b=0.5$ and $k=9$.
Finally, note that this procedure only works if the training data is approximately translationally invariant.

\begin{figure}[htb]
    \centering
    \includegraphics[width=0.8\linewidth]{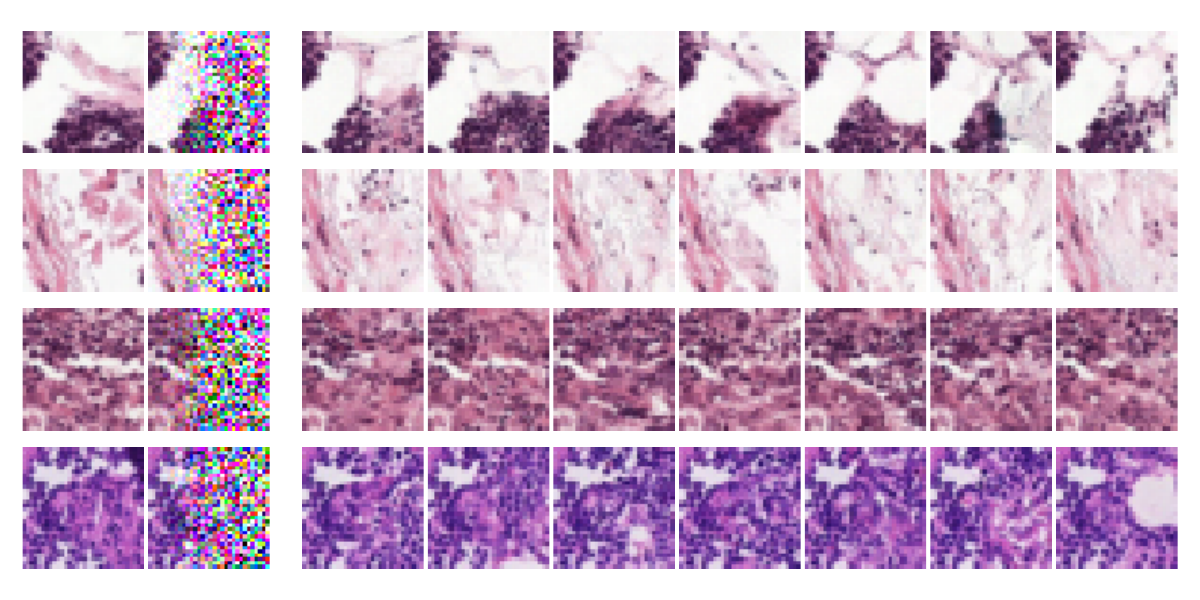}
    \vspace{-10pt}
   \caption{\footnotesize Reconstruction of images from the test set (left column) partially noised to $t=0.5$ (second column) using a sequential schedule in real space as described in \S \ref{sec:column}. The different reconstructions shown on the right differ by random key used.}
    \label{fig:pcam-reconstruct}
\end{figure}

\section{Haar wavelets}

The 2D Haar wavelet transform decomposes an image $X \in \mathbb{R}^{N \times N \times C}$ into low- and high-frequency components across multiple scales.
We apply the same wavelet transform to each color channel $c = 1, \dots, C$ (for us $C=3$) separately.
Therefore, to ease the notation we will suppress the color index in what follows. The wavelet transform
at level  $n$ is defined recursively as follows:

\paragraph*{1. Row Transformation}
   Apply the 1D Haar transform along the rows:
   \[
   \begin{aligned}
   L_{i,k}^{(n)} &= \tfrac{1}{\sqrt{2}} \left( X_{2i,k}^{(n-1)} + X_{2i+1,k}^{(n-1)} \right), \\
   H_{i,k}^{(n)} &= \tfrac{1}{\sqrt{2}} \left( X_{2i,k}^{(n-1)} - X_{2i+1,k}^{(n-1)} \right),
   \end{aligned}
   \]
   where \( i = 0, \dots, \tfrac{N}{2^{n}} - 1 \) and \( k = 0, \dots, N - 1 \).

\paragraph*{2. Column Transforms}
Apply the 1D Haar transform along the columns to the results of the row transformation:
\begin{equation}
   \begin{aligned}
   LL_{i,j}^{(n)} &= \tfrac{1}{\sqrt{2}} \left( L_{i,2j}^{(n)} + L_{i,2j+1}^{(n)} \right), \\
   LH_{i,j}^{(n)} &= \tfrac{1}{\sqrt{2}} \left( L_{i,2j}^{(n)} - L_{i,2j+1}^{(n)} \right), \\
   HL_{i,j}^{(n)} &= \tfrac{1}{\sqrt{2}} \left( H_{i,2j}^{(n)} + H_{i,2j+1}^{(n)} \right), \\
   HH_{i,j}^{(n)} &= \tfrac{1}{\sqrt{2}} \left( H_{i,2j}^{(n)} - H_{i,2j+1}^{(n)} \right),
   \end{aligned}
\end{equation}
where \( j = 0, \dots, \tfrac{N}{2^{n}} - 1 \).

\paragraph*{3. High-Frequency Component:}
Stack the high-frequency sub-bands into a single high-frequency array at level \( n \):
\begin{equation}
    HF^{(n)} = \text{concat}\left( LH^{(n)},\, HL^{(n)},\, HH^{(n)} \right) \,.
\end{equation}

\paragraph*{4. Recursive Decomposition}

The low-frequency component $LL^{(n)}$ becomes the input for the next level:
\begin{equation}
    X^{(n)} = LL^{(n)} \,.
\end{equation}

At each level, the transform produces one array of low-frequency components $LL^{(n)}$ and one array of high-frequency components $HF^{(n)}$. This process can be recursively applied up to a desired depth ${\cal N}$, resulting in a hierarchical decomposition of the image.

After level ${\cal N}$, the original image is represented by one lowest-frequency array and ${\cal N}$ higher-frequency arrays.

For example, for level 3 one obtains three high-frequency arrays $HF^{(1)}$, $HF^{(2)}$, $HF^{(3)}$,
and one coarse array $ LL^{(3)}$.
To accommodate images with multiple color channels $C$, the transform is applied independently to each channel, and the resulting components are concatenated along the channel dimension.
The factors of $\sqrt{2}$ make sure that the transform is an orthogonal transformation, whose inverse can be computed analogously.

\end{document}

%% file: math_commands.tex
\usepackage{amsmath,amsfonts,bm}

\def\eqref#1{equation~\ref{#1}}

\def\1{\bm{1}}

\DeclareMathAlphabet{\mathsfit}{\encodingdefault}{\sfdefault}{m}{sl}
\SetMathAlphabet{\mathsfit}{bold}{\encodingdefault}{\sfdefault}{bx}{n}



%% file: distribution-path.tex
    
    
    
    
    


\begin{tikzpicture}[
    node distance=0.0cm and 0.0cm,
    every node/.style={font=\small},
    dot/.style={circle, fill=black, minimum size=6pt, inner sep=0pt},
    line/.style={black, line width=1pt},
    dashedline/.style={black, dashed, line width=1pt},
    label/.style={font=\small, align=center}
  ]
    
    \coordinate (B) at (0,0);
    \coordinate (A) at (3,2); 

    \begin{scope}
        \clip (A) to[out=180, in=70] (B) -- (B) to[out=10, in=260] (A) -- cycle;
        \fill[gray, opacity=0.2] (-1,-1) rectangle (4,3);
    \end{scope}

    \draw[line] (A) -- (B);
    
    \draw[dashedline] (A) to[out=180, in=70] (B); 
    \draw[dashedline] (A) to[out=260, in=10] (B); 
    \draw[dashedline] (A) to[out=240, in=20] (B); 
    
    \node[dot] (dotA) at (A) {};
    \node[dot] (dotB) at (B) {};

    \node[label] at ($(A) + (-2, 0.2)$) (labelA) {$p_{\rm data}$; $\boldsymbol{\alpha} = 1$, $\boldsymbol{\gamma} \rightarrow -\infty$};
    \node[label] at ($(B) + (2, -0.2)$) (labelB1) {$p_{\rm prior}$; $\boldsymbol{\alpha} = 0$, $\boldsymbol{\gamma} \rightarrow \infty$};

\end{tikzpicture}